\documentclass[]{malab}

\usepackage{array}
\usepackage{makecell}
\usepackage{colortbl}
\usepackage{amsmath}
\usepackage{enumitem}
\definecolor{PATTBlue}{RGB}{67,101,160}
\definecolor{PATTBlueLight}{RGB}{231,238,250}
\definecolor{PATTGreen}{RGB}{62,132,92}
\definecolor{PATTGreenLight}{RGB}{231,244,233}
\definecolor{PATTOrange}{RGB}{213,119,54}
\definecolor{PATTOrangeLight}{RGB}{250,239,226}
\definecolor{PATTGray}{RGB}{238,238,240}
\definecolor{PATTTextGray}{RGB}{92,92,98}

\newcommand{\method}{PATT\xspace}

\newcommand{\sg}{\operatorname{sg}}
\DeclareMathOperator{\IoU}{IoU}
\DeclareMathOperator{\clip}{clip}

\hypersetup{
    colorlinks=true,
    linkcolor=PATTBlue,
    citecolor=PATTBlue,
    urlcolor=malabblue
}

\title{Learning to Track from Privileged\\ Target Appearances}
\shorttitle{Learning to Track from Privileged Target Appearances}

\author[1]{Xin Chen}
\author[2]{Jiao Xu}
\author[2]{Dong Wang}
\author[2]{Huchuan Lu}
\author[1]{Kede Ma}
\affiliation[1]{City University of Hong Kong}
\affiliation[2]{Dalian University of Technology}
\authoremails{\email{xche32@cityu.edu.hk}, \email{xjmmcome@mail.dlut.edu.cn}, \email{wdice@dlut.edu.cn}, \email{lhchuan@dlut.edu.cn}}
\correspondence{\email{kede.ma@cityu.edu.hk}}
\date{September 2, 2026}

\projectpage{\url{ https://github.com/Multimedia-Analytics-Laboratory/PATT}}

\abstract{
Target templates define what a visual tracker searches for, yet the templates available at inference trade off \textit{localization certainty} with \textit{appearance freshness}: the initial ground-truth template is exact but becomes stale, whereas recent templates better reflect the current appearance but are cropped from uncertain predictions.
We quantify this bottleneck with a non-deployable oracle that supplies an exact current-frame target crop,
improving AUC on LaSOT by $15.2$ percentage points. This gap reveals a training-only opportunity:
frame-level ground truths provide exact current- and future-frame target crops,
although such crops are unavailable at deployment. We introduce Privileged Appearance Transfer for Tracking (\textbf{\method}), a teacher-student training framework that transfers these privileged appearances to a deployable tracker through multi-level representation prediction.
The privileged teacher observes exact target crops from past, current, and future frames, whereas the student receives only past-frame templates and learns to predict the teacher's search representations.
To avoid transferring unreliable teacher signals, \method weights this transfer by the teacher’s relative localization advantage over the student and its absolute localization accuracy.
After training, the teacher, latent predictor, reliability weights, and privileged crops are removed, leaving standard student-only inference.
Across seven benchmarks at two model scales, \method achieves consistent gains under both long- and short-term tracking protocols.
}

\begin{document}

\maketitle

\section{Introduction}
\label{sec:introduction}

Tracking is fundamentally a problem of maintaining object correspondence through time.
Humans can preserve object identity across motion even in the presence of visually identical distractors~\citep{pylyshyn1988tracking}, while object-file theory explains this ability as the integration of successive visual states into temporary object representations~\citep{kahneman1992objectfiles}.
Visual object tracking poses a computational analogue: given a target in the first frame, localize the same object as its position and appearance evolve~\citep{lucas1981iterative,OTB2013}.
Modern trackers typically encode target identity through image templates, which specify \emph{what} to track and guide localization in each new search region. This template-search paradigm underlies Siamese trackers~\citep{SiameseFC,SiameseRPN}, Transformer trackers~\citep{transt,Stark,mixformer,ostrack}, autoregressive trackers~\citep{seqtrack,artrack}, and unified tracking models~\citep{OneTracker,sutrack}. Despite their architectural diversity, these methods share a common dependence on whether the available templates remain discriminative for the target's current state.

This dependence is complicated by the fact that the target representation needed for correspondence changes over time.
The initial template is extracted from the given ground-truth bounding box and is therefore \textit{exact} and \textit{localization-certain}, but its appearance may become obsolete under deformation, rotation, illumination change, scale variation, or occlusion.
A template acquired from a recent frame can better reflect the target's current appearance, yet its crop must be derived from the tracker's own predicted box.
When this prediction contains substantial background or drifts toward a distractor, template updates can reinforce the resulting error.
We refer to this limitation as the \textbf{inference-time template acquisition bottleneck}: at deployment, no reliably available template jointly provides localization certainty and appearance freshness.

Existing trackers mitigate this bottleneck by extracting additional target evidence from the inference history.
Online discriminative learning~\citep{DiMP,ToMP}, confidence-based template updates~\citep{Stark,mixformer}, temporal token propagation~\citep{odtrack,atctrack,dtptrack,relo2026}, and autoregressive state modeling~\citep{seqtrack,artrack,artrackv2} all seek to preserve reliable target information as a video unfolds. However,
they cannot remove the underlying information constraint: apart from the initial template, every newly acquired target crop is derived from a model prediction and may therefore inherit its localization errors.
Without ground-truth supervision, an inference-time update rule cannot verify whether a new crop precisely depicts the target.

\begin{figure*}[t]
\centering
\includegraphics[width=.85\linewidth]{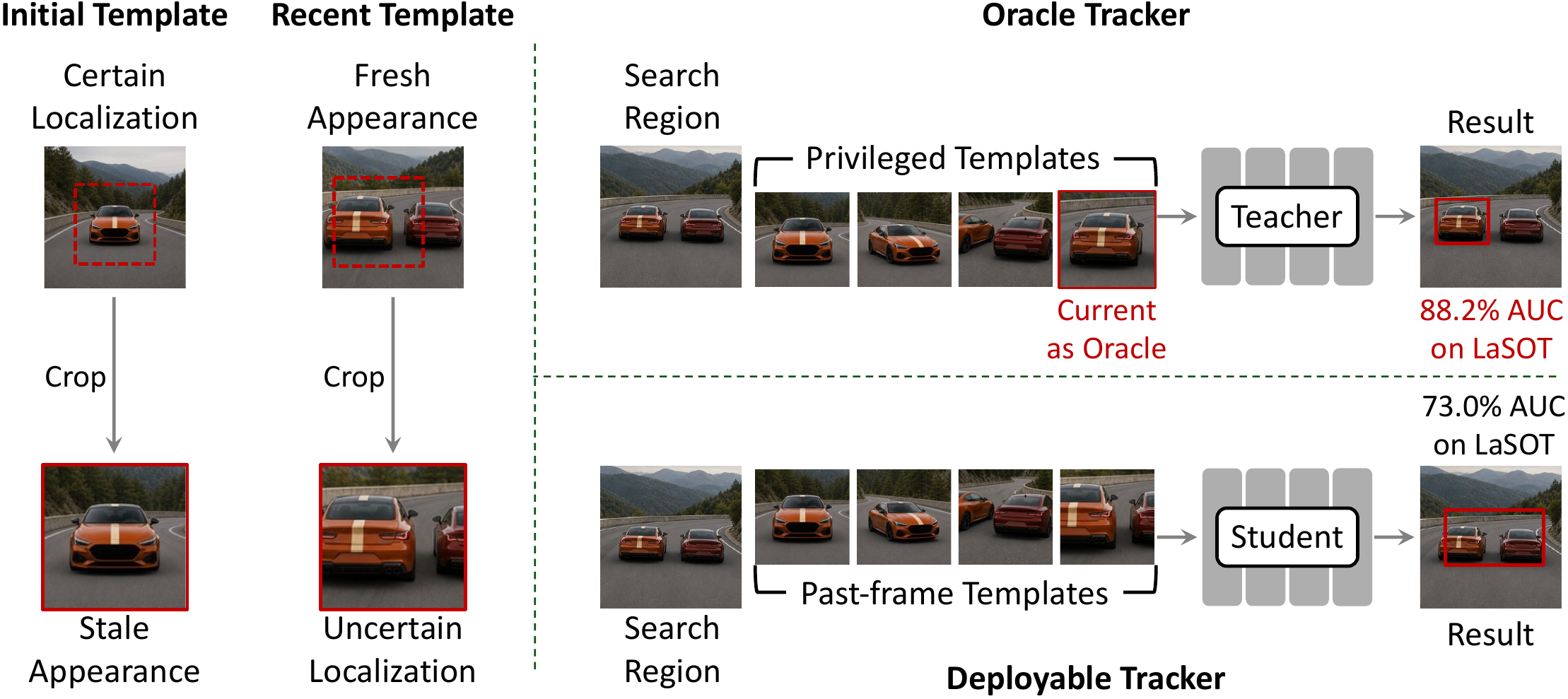}
\caption{\textbf{Oracle study on LaSOT.}
Standard trackers face an inference-time template-acquisition bottleneck: the initial template is accurately localized but becomes stale, whereas recent templates are fresher but cropped from uncertain predictions.
A non-deployable oracle removes this  bottleneck by using the ground-truth bounding box to supply an exact current-frame target crop.
On LaSOT, AUC increases from $73.0\%$ to $88.2\%$, a substantial gain of $15.2$ percentage points.}
\label{fig:motivation}
\end{figure*}

Training exposes precisely the information missing at inference.
Frame-level ground-truth bounding boxes provide exact target crops not only before the search frame, but also from the current and future frames.
To quantify the value of this information, we evaluate a \textit{non-deployable} oracle that gives the tracker an exact current-frame target crop.
As shown in \Cref{fig:motivation}, this single intervention raises the area under the success curve (AUC) on LaSOT~\citep{LaSOT} from $73.0\%$ to $88.2\%$, a substantial gain of $15.2$ percentage points.
This oracle gap suggests that exact, up-to-date target appearances remain a major performance-limiting factor. Yet standard tracking objectives use the current ground-truth bounding box primarily for localization supervision and past target crops as causal templates; they barely exploit exact current and future target appearances as training-only inputs.

Motivated by this oracle gap, we introduce \textbf{P}rivileged \textbf{A}ppearance \textbf{T}ransfer for \textbf{T}racking (\textbf{\method}), a teacher-student training framework that turns training-only target appearances into supervision for a deployable tracker.
The student uses four past-frame target templates, whereas the privileged teacher replaces two of them
with exact target crops from the current and a future frame. Apart from this difference in template source, the two branches share the same architecture, backbone capacity, and four-slot template budget, making privileged target appearances the teacher's principal information advantage.

Rather than imitate the teacher's final box predictions, for which ground-truth bounding boxes already provide direct supervision,
\method transfers privileged information through multi-level search representations.
A latent predictor maps the student's multi-level search representations to their counterparts produced by the privileged teacher, which is maintained as an exponential moving average (EMA) of the student
to provide stable targets. However, even with privileged inputs, some teacher predictions may still be \textit{unreliable}. \method therefore weights the transfer using two complementary criteria: the teacher’s relative localization advantage over the student and its absolute localization accuracy. At deployment, the teacher, latent predictor, reliability weights, and privileged crops are discarded,
leaving the standard deployable tracker unchanged.

Experiments across seven benchmarks demonstrate that \method achieves consistent gains across Base and Large model scales, long- and short-term tracking protocols, and datasets outside the training distribution.
In particular, \method-L achieves $76.7\%$ AUC on LaSOT, $56.7\%$ AUC on LaSOT$_\mathrm{ext}$~\citep{lasot_journal}, and $64.8\%$ AUC on TNL2K~\citep{TNL2K}.
These results show that privileged target appearances can improve the representations of a deployable tracker without additional inference components.

In summary, our contributions are as follows:
\begin{itemize}[leftmargin=1.25em,itemsep=0.25em,topsep=0.3em,parsep=0pt]
\item We identify the inference-time template acquisition bottleneck in visual tracking and quantify its impact with a non-deployable oracle that improves LaSOT AUC by $15.2$ percentage points.
\item We introduce \method, a teacher-student framework that transfers privileged target appearances to a deployable tracker through multi-level representation prediction and reliability weighting.
\item We demonstrate consistent improvements across seven benchmarks, two model scales, and multiple evaluation regimes, showing the effectiveness of training-only target appearances.
\end{itemize}

\section{Related Work}
\label{sec:related}

In this section, we position \method relative to two lines of work: template-based visual tracking and learning with privileged information.
The former exposes the causal constraint at deployment, whereas the latter frames the use of training-only target crops as privileged supervision for a deployable tracker.

\subsection{Template-based Visual Tracking}

Template conditioning is a central paradigm in visual tracking. 
Siamese trackers instantiate this idea by comparing a target template with a search region, as in SiamFC~\citep{SiameseFC}, and by adding region proposals, as in SiamRPN~\citep{SiameseRPN}.
 Transformer trackers further strengthen template-search interaction through explicit feature fusion, as in TransT~\citep{transt}, STARK~\citep{Stark}, and MixFormer~\citep{mixformer}, or through joint token processing in a unified encoder, as in OSTrack~\citep{ostrack}. Recent work extends this paradigm with larger pretrained backbones~\citep{LoRAT}, autoregressive target prediction~\citep{seqtrack,artrack,artrackv2}, and architectures that unify tracking with other forms of target specification~\citep{OneTracker,sutrack}.

A complementary line of research concerns how target evidence can be maintained after initialization. Online discriminative methods such as ATOM \citep{ATOM}, DiMP~\citep{DiMP}, and ToMP~\citep{ToMP} update target models using newly collected positive and negative samples. Template-update methods such as STARK~\citep{Stark} and MixFormer~\citep{mixformer} refresh stored templates from confident predictions, while KeepTrack~\citep{keeptrack} associates candidate states across frames to reduce ambiguity. Other approaches, including ODTrack~\citep{odtrack}, autoregressive trackers~\citep{seqtrack,artrack}, and RELO~\citep{relo2026} condition localization on previous predictions or latent target states. These mechanisms improve temporal adaptation, but they remain constrained by causal information available at deployment: except for the initial target crop, newly acquired target evidence is derived from model predictions and is therefore susceptible to localization error. \method instead studies whether a deployable tracker can benefit from current- and future-frame target crops extracted from ground-truth bounding boxes during training, without changing its inference-time inputs or computation.

\subsection{Learning with Privileged Information}

Learning with privileged information formalizes settings in which auxiliary variables
are available during training but absent at deployment~\citep{vapnik2009new}.
The central challenge is to transfer this training-only information to a model
that relies only on deployable inputs.
Knowledge distillation provides a natural mechanism for such transfer:
a student learns from targets produced by a teacher~\citep{hinton2015distilling}.
Generalized distillation connects the two ideas by using privileged variables
to construct an informative teacher whose predictions supervise a deployable student
~\citep{lopezpaz2016unifying}.

Teacher-student paradigms differ in what gives the teacher an advantage
and what signal is transferred to the student.
In conventional distillation, the teacher's advantage typically comes from
greater model capacity or an ensemble~\citep{hinton2015distilling}.
In privileged-information settings, the advantage instead comes from
training-only annotations or modalities, such as semantic object attributes,
bounding boxes~\citep{motiian2016privileged}, or depth~\citep{hoffman2016hallucination}.
The transferred signal may be an output distribution~\citep{hinton2015distilling}
or an intermediate representation~\citep{romero2015fitnets}.
A separate design choice is teacher stability.
Mean Teacher~\citep{tarvainen2017mean} maintains the teacher as an EMA of the student to provide stable consistency targets,
but parameter averaging alone does not create an information advantage.

Teacher-student transfer has also been explored in Siamese tracking
~\citep{DSTrpn}.
\method differs from these settings by keeping the teacher and student at the same
architecture and capacity: the teacher's advantage comes solely from exact
current- and future-frame target crops available during training, and the
transferred signal is its intermediate search representations.

\section{Privileged Appearance Transfer for Tracking}
\label{sec:method}

\begin{figure}[t]
\centering
\includegraphics[width=0.95\linewidth]{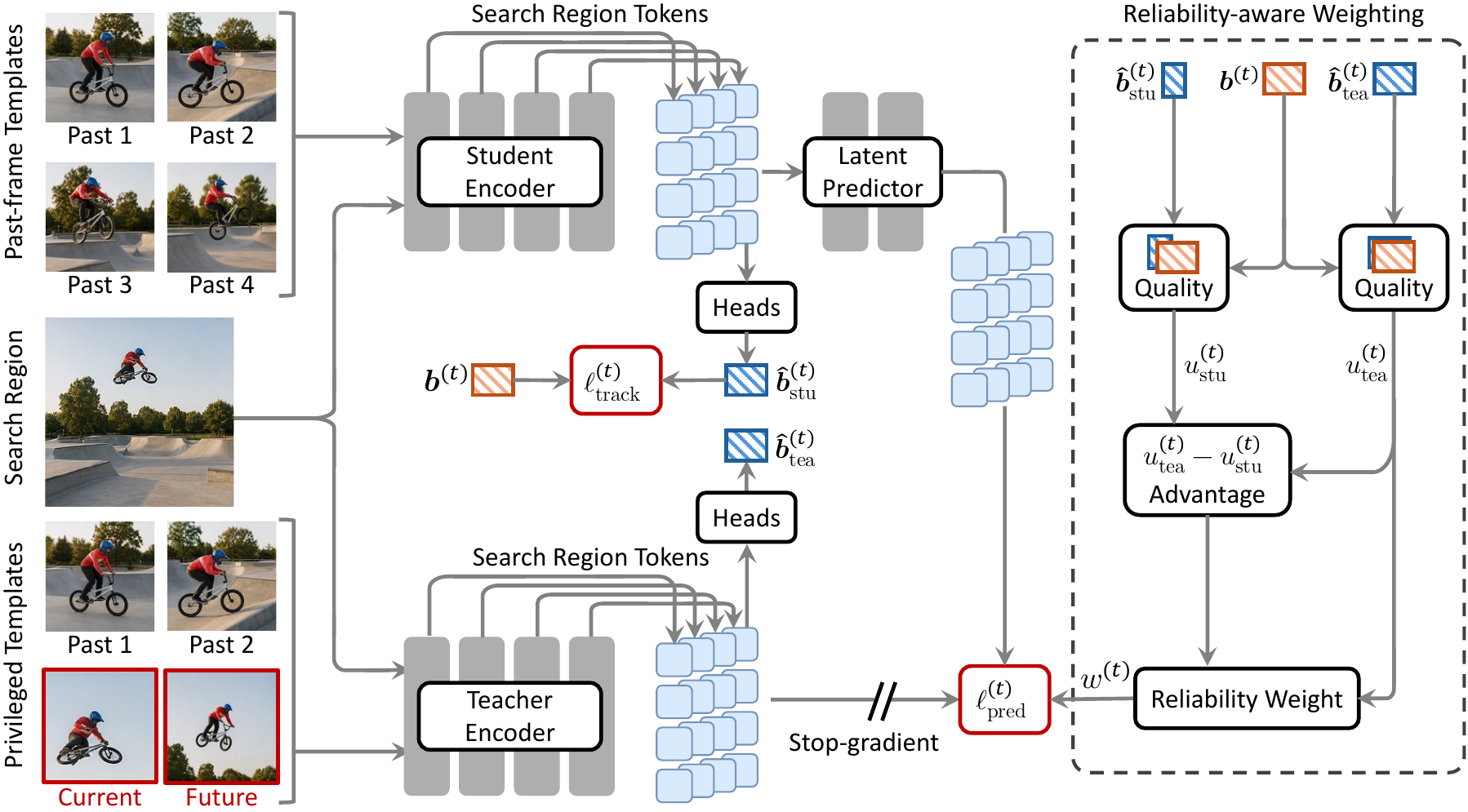}
\caption{\textbf{Overview of \method.}
The student and privileged teacher process the same search crop with the same tracker architecture, model capacity, and four-slot template budget.
The student uses only past-frame templates, whereas the teacher replaces two template slots with exact current- and future-frame target crops extracted from ground-truth bounding boxes.
A latent predictor maps the student's multi-level search representations to the teacher's stop-gradient representations, defining the representation-prediction loss.
A reliability weight modulates this loss by combining the teacher's relative localization advantage over the student and its absolute localization quality.
At deployment, the teacher, predictor, reliability computation, and privileged
crops are discarded, leaving only the student tracker.}
\label{fig:framework}
\end{figure}
\vspace{-3mm}

In this section, we present \method, a training framework that transfers privileged target-appearance information to a deployable tracker, while leaving its inference procedure unchanged.
As shown in \Cref{fig:framework}, the teacher shares the student's tracker architecture and template budget, but replaces two past-frame templates with current- and future-frame target crops extracted from ground-truth bounding boxes.
A latent predictor maps the student's multi-level search representations to the corresponding teacher representations, and a sample-wise reliability weight modulates this transfer.
After training, all privileged inputs and auxiliary modules are discarded, leaving only the standard student tracker.

\subsection{Template-Based Tracking Formulation}

\paragraph{Student template construction.}
We first formulate the causal template input used by the student tracker. We represent a $T$-frame training video $\mathbf{I}$ and its ground-truth bounding boxes $\mathbf{B}$ as ordered sequences:
\begin{equation}
\mathbf{I}
=
\left(\mathbf{I}^{(1)},\ldots,\mathbf{I}^{(T)}\right),
\qquad
\mathbf{B}
=
\left(\bm{b}^{(1)},\ldots,\bm{b}^{(T)}\right).
\end{equation}
Let $\operatorname{crop}(\cdot,\cdot)$ denote the template-extraction operator of the underlying tracker. The target template extracted from frame $i$ is
\begin{equation}
\bm{x}^{(i)}
=
\operatorname{crop}\!\left(\mathbf{I}^{(i)},\bm{b}^{(i)}\right).
\end{equation}
For search frame $t$, let $r_k$ index the past frame used to extract the $k$-th student template, where $1\leq r_1<\cdots< r_K<t$ and $K\ge 2$. The student's ordered template input is
\begin{equation}
\mathbf{X}_{\mathrm{stu}}^{(t)}
=
\left(\bm{x}^{(r_1)},\ldots,\bm{x}^{(r_K)}\right).
\label{eq:student_templates}
\end{equation}
During training, these past-frame templates are extracted using ground-truth bounding boxes. At inference, however, ground-truth is available only for the initial template; any later template must be cropped from a predicted target box.

\paragraph{Student forward pass.}
Let $\bm{s}^{(t)}$ denote the search crop sampled from $\mathbf{I}^{(t)}$ around the ground-truth box $\bm{b}^{(t)}$.
We use $\mathcal{L}$ for the set of encoder layers selected for representation transfer, with $L=|\mathcal{L}|$, and arrange their outputs in increasing order of encoder depth when stacked. Given the student templates and search crop, the tracker predicts a target box and produces search representations at the selected layers:
\begin{equation}
\left(
\left\{\mathbf{Z}_{\mathrm{stu},l}^{(t)}\right\}_{l\in\mathcal{L}},
\hat{\bm{b}}_{\mathrm{stu}}^{(t)}
\right)
=
\bm{f}_{\bm{\theta}}
\!\left(\mathbf{X}_{\mathrm{stu}}^{(t)},\bm{s}^{(t)}\right),
\label{eq:student_forward}
\end{equation}
where $\mathbf{Z}_{\mathrm{stu},l}^{(t)}\in\mathbb{R}^{N\times D}$ contains the $D$-dimensional representations of the $N$ student search tokens at layer $l$.

\paragraph{Standard tracking objective.}
The student parameters $\bm{\theta}$ are optimized using the tracker's standard localization objective:
\begin{equation}
\ell_{\mathrm{track}}^{(t)}(\bm{\theta})
=
\ell_{\mathrm{loc}}
\left(\bm{b}^{(t)}, \hat{\bm{b}}_{\mathrm{stu}}^{(t)}\right),
\label{eq:tracking_loss}
\end{equation}
where the dependence on $\bm{\theta}$ follows from the forward pass in \Cref{eq:student_forward}. Additional head outputs and tracker-specific targets are left implicit. This objective provides direct supervision for localization, while the privileged teacher introduced in the next subsection supplies an additional representation-level signal at intermediate encoder layers.

\subsection{Multi-Level Representation Prediction}

We now construct a privileged teacher that accesses training-only information while keeping the tracker architecture and template budget fixed. The teacher provides multi-level search representations as targets for the student, and its box prediction is used only to measure the reliability of this transfer.

\paragraph{Matched privileged templates.}
The oracle study in \Cref{fig:motivation} demonstrates the potential value of privileged target information: supplying the tracker with an exact current-frame target crop raises LaSOT AUC from $73.0\%$ to $88.2\%$. This $15.2$-point gain motivates using ground-truth bounding boxes during training
to construct privileged target crops that are unavailable to a deployable tracker.

We instantiate this privilege with two teacher-only templates. For search frame $t$, the current-frame crop $\bm{x}^{(t)}$ depicts the target in the same frame as the search crop $\bm{s}^{(t)}$. A future-frame crop $\bm{x}^{(t')}$, with $t<t'\leq T$, provides a later target appearance that may capture subsequent variation. Neither crop is a valid deployment-time input for localizing the target at frame $t$: the current crop requires the unknown target box at frame $t$, and the future crop additionally requires access to a later frame.

Rather than increasing the template budget, we form the teacher input by replacing the student's final two past-frame templates in \Cref{eq:student_templates} with the privileged current- and future-frame crops:
\begin{equation}
\mathbf{X}_{\mathrm{tea}}^{(t)}
=
\left(\bm{x}^{(r_1)},\ldots,\bm{x}^{(r_{K-2})},
 \bm{x}^{(t)},\bm{x}^{(t')}\right).
\label{eq:matched_layout}
\end{equation}
The student and teacher otherwise share the same search crop, tracker architecture, model capacity, and template budget.

\paragraph{Teacher representation prediction.}
Given the privileged template tuple and the shared search crop, the teacher produces search representations at the selected encoder layers and predicts a target box:
\begin{equation}
\left(
\left\{\mathbf{Z}_{\mathrm{tea},l}^{(t)}\right\}_{l\in\mathcal{L}},
\hat{\bm{b}}_{\mathrm{tea}}^{(t)}
\right)
=
\bm{f}_{\bar{\bm{\theta}}}
\!\left(\mathbf{X}_{\mathrm{tea}}^{(t)},\bm{s}^{(t)}\right),
\label{eq:teacher_forward}
\end{equation}
where $\bar{\bm{\theta}}$ denotes the teacher parameters.

We transfer the teacher's intermediate search representations rather than imitating its predicted box. This is because box-level imitation would add another localization target on top of the direct ground-truth supervision in \Cref{eq:tracking_loss}, while exposing only the teacher's final localization outcome. Intermediate representations instead capture the token-level evidence shaped by the privileged current- and future-frame templates, providing
complementary supervision for the student's encoder.

Following multi-level feature prediction~\citep{vjepa21}, we apply LayerNorm separately to the outputs of the $L$ selected encoder layers in each branch, and concatenate them from shallow to deep along the feature dimension. This yields student and teacher representations \(\mathbf Z_{\mathrm{stu}}^{(t)},\mathbf Z_{\mathrm{tea}}^{(t)}\in\mathbb R^{N\times(D\times L)}\). Because privileged templates change how the teacher organizes search evidence, we do not force the student representations  $\mathbf{Z}_{\mathrm{stu}}^{(t)}$ to match the teacher representations $\mathbf{Z}_{\mathrm{tea}}^{(t)}$ directly. Instead, we learn a Transformer predictor $\bm{g}_{\bm \phi}:\mathbb{R}^{N\times (D\times L)}\mapsto\mathbb{R}^{N\times (D\times L)}$ that maps $\mathbf{Z}_{\mathrm{stu}}^{(t)}$ to $\mathbf{Z}_{\mathrm{tea}}^{(t)}$. 

\paragraph{Prediction objective and teacher update.}
During each student update, the teacher target is detached with the stop-gradient operator $\sg[\cdot]$, which is the identity in the forward pass and has zero derivative in the backward pass. Consequently, the representation-prediction loss is
\begin{equation}
\ell_{\mathrm{pred}}^{(t)}(\bm \theta, \bm \phi;\bar{\bm \theta})
=
\frac{1}{NDL}
\left\|
\bm{g}_{\bm \phi}\!\left(\mathbf{Z}_{\mathrm{stu}}^{(t)}\right)
-
\sg\!\left[\mathbf{Z}_{\mathrm{tea}}^{(t)}\right]
\right\|_{1,1},
\label{eq:predictive_loss}
\end{equation}
 where $\|\cdot\|_{1,1}$ sums absolute values over all matrix entries. The denominator averages the prediction error over search tokens, feature channels, and selected encoder layers. Gradients thus update the student and predictor, but not the teacher. Instead, we adjust the teacher parameters as an EMA of the student parameters:
\begin{equation}
\bar{\bm{\theta}}
\leftarrow
\rho\bar{\bm{\theta}}+(1-\rho)\bm{\theta},
\qquad \rho\in[0,1).
\label{eq:ema_update}
\end{equation}
The privileged templates and EMA play distinct roles: the templates determine what additional information the teacher observes, while EMA stabilizes the teacher's parameter trajectory and representation targets.

\subsection{Reliability Weighting}

Privileged inputs make the teacher more informative, but they do not necessarily make every teacher representation equally reliable as a prediction target. For a given sample, the teacher may still localize poorly, or it may be no better than the student. We therefore use localization quality as a training-time proxy for transfer reliability, weighting representation prediction by the teacher's relative advantage over the student and its absolute localization accuracy.

\paragraph{Reliability weight.}
For frame $t$, we measure the localization quality of each branch by its intersection over union (IoU) with the ground-truth bounding box:
\[
u_{\mathrm{stu}}^{(t)}
=
\IoU\!\left(\hat{\bm{b}}_{\mathrm{stu}}^{(t)},\bm{b}^{(t)}\right),
\qquad
u_{\mathrm{tea}}^{(t)}
=
\IoU\!\left(\hat{\bm{b}}_{\mathrm{tea}}^{(t)},\bm{b}^{(t)}\right).
\]
The reliability weight combines a relative advantage term with an absolute quality term:
\begin{equation}
w^{(t)}
=
\sg\!\left[
\underbrace{
\clip\!\left(
\frac{u_{\mathrm{tea}}^{(t)}-u_{\mathrm{stu}}^{(t)}}{\tau},0,1
\right)
}_{\text{advantage}}
\;
\underbrace{
\!\left(u_{\mathrm{tea}}^{(t)}\right)^{\gamma}
}_{\text{quality}}
\right],
\qquad
\tau>0,\ \gamma\geq0,
\quad
w^{(t)}\in[0,1].
\label{eq:reliability}
\end{equation}
The clipped advantage term suppresses transfer when the teacher is no more accurate than the student, increases linearly with the teacher's positive IoU margin, and saturates once this margin reaches $\tau$. However, a positive margin alone is insufficient: the teacher may outperform the student even when both predictions are poor. The quality term therefore discounts such cases according to the teacher's absolute IoU. With the convention $(u_{\mathrm{tea}}^{(t)})^0=1$, setting $\gamma=0$ recovers advantage-only weighting.

\paragraph{Overall training objective.}
We define the complete per-sample  training objective as
\begin{equation}
\ell^{(t)}(\bm \theta,\bm \phi)
=
\ell_{\mathrm{track}}^{(t)}(\bm \theta)
+
\lambda w^{(t)}\ell_{\mathrm{pred}}^{(t)}(\bm \theta,\bm \phi),
\qquad
\lambda\geq0,
\label{eq:full_loss}
\end{equation}
where $\lambda$ controls the strength of privileged representation prediction relative to standard tracking supervision. The tracking loss is applied to every sample, while $w^{(t)}$ adjusts the representation-prediction loss according to the teacher's reliability. The stop-gradient in \Cref{eq:reliability} treats $w^{(t)}$ as a fixed sample-dependent coefficient during backpropagation. Without this detachment, the student could reduce the weighted prediction loss $w^{(t)}\ell_{\mathrm{pred}}^{(t)}$ by changing its localization output to lower $w^{(t)}$, rather than by predicting the teacher representation more accurately. Detaching $w^{(t)}$ prevents this shortcut and restricts the reliability weight to controlling the sample-wise strength of representation supervision.

\section{Experiments}
\label{sec:experiments}

In this section, we evaluate whether privileged appearance transfer improves a deployable tracker that uses only the student branch at inference. We first specify the training and inference protocols, then compare two model scales across seven benchmarks, and finally use controlled ablations to isolate the effects of privileged appearances, multi-level prediction, and reliability weighting.

\subsection{Experimental Setups}

\begin{table}[t]
\centering
\caption{\textbf{Model configurations and inference cost.}
Parameter counts and floating-point operations (FLOPs) outside parentheses are for  the deployed tracker, while  values in parentheses indicate the additional training-only cost of the representation predictor.
Speed is reported in frames per second (FPS) on a single NVIDIA RTX 4090 GPU.}
\label{tab:model_variants}
\resizebox{1\linewidth}{!}{
\setlength{\tabcolsep}{12.5mm}
\begin{tabular}{lccc}
\toprule
Model & \# Params (M) & FLOPs (G) & Speed (FPS) \\
\midrule
\method-B256 & $70\,(+23)$  & $39\,(+6)$  & $50$ \\
\method-L256 & $247\,(+25)$ & $131\,(+6)$ & $32$ \\
\bottomrule
\end{tabular}}
\end{table}

\paragraph{Model variants.}
We instantiate \method in the one-stream Transformer tracking framework of OSTrack~\citep{ostrack}.
\method-B256 and \method-L256 use the Base and Large Fast-iTPN encoders~\citep{fastitpn}, respectively. \Cref{tab:model_variants} reports their deployed costs.
Both variants use a $256\times256$ search crop, $128\times128$ templates, a patch stride of $16$, $K=4$ template slots, and the center-based prediction head of OSTrack.
The training-only representation predictor comprises $12$ Transformer layers with a hidden dimension of $384$ and $12$ attention heads.

\paragraph{Training protocol.}
We train on the training splits of COCO~\citep{COCO}, LaSOT~\citep{LaSOT}, GOT-10k~\citep{GOT10K}, TrackingNet~\citep{trackingnet}, and VastTrack~\citep{vasttrack}.
For each training example, we sample ordered frames $r_1<r_2<r_3<r_4<t<t'$ from one video.
The student uses target crops from frames $r_1,\ldots,r_4$ as templates and the crop from frame $t$ as the search input. The teacher uses the same search crop but replaces the final two past-frame templates with ground-truth target crops from frames $t$ and $t'$.
We obtain template and search crops by expanding their ground-truth bounding boxes by factors of $2$ and $4$, respectively, and further apply translation and scale jitter to the search crop.
We train with AdamW for \(10\) epochs using a minibatch size of \(64\) and an initial learning rate of \(10^{-4}\), decayed by a factor of \(10\) after epoch \(8\).
We predict representations from $L=4$ evenly spaced encoder layers and update the teacher by EMA with \(\rho=0.99925\).
The loss hyperparameters are $\lambda=3$, $\tau=0.1$, and $\gamma=0.5$.

\paragraph{Inference protocol.}
At test time, only the student $\bm{f}_{\bm{\theta}}$ is retained; the teacher, predictor, reliability computation, and privileged templates are removed.
Following the template-update protocol of the underlying tracker~\citep{Stark}, the student keeps the ground-truth first-frame template and fills the remaining three slots with dynamic templates cropped from its own predictions.
Dynamic templates are refreshed every $25$ frames when prediction confidence exceeds $0.80$.

\paragraph{Evaluation protocol.}
We evaluate on seven benchmarks covering different tracking regimes. LaSOT~\citep{LaSOT} and LaSOT$_\mathrm{ext}$~\citep{lasot_journal} evaluate long-horizon appearance change and cross-category generalization. TrackingNet~\citep{trackingnet} and GOT-10k~\citep{GOT10K} test large-scale short-term tracking. TNL2K~\citep{TNL2K}, NFS~\citep{NFS}, and UAV123~\citep{UAV} assess transfer beyond the training datasets.
We use the area under the success curve (AUC) as the primary metric. LaSOT, LaSOT$_\mathrm{ext}$, and TrackingNet additionally report precision ($P$) and normalized precision ($P_{\mathrm{Norm}}$), while GOT-10k reports average overlap (AO) and success rates at IoU thresholds $0.5$ and $0.75$.

\subsection{Main Results}

\begin{table*}[t!]
  \centering
  \caption{\textbf{Comparison on four large-scale tracking benchmarks.}
  Results are grouped by model scale, with base-scale trackers in the upper block and large-scale trackers in the lower block.
  AUC, $P$, $P_{\mathrm{Norm}}$, AO, and SR denote area under the success curve, precision, normalized precision, average overlap, and success rate.
  Within each block, \textbf{bold} and \underline{underlined} values indicate the best and second-best results, respectively.}
\label{tab:main}
\resizebox{1\linewidth}{!}{
  \setlength{\tabcolsep}{0.5mm}
  \small
  \begin{tabular}{l ccc c ccc c ccc c ccc}
    \toprule
    \multirow{2}*{Model} & \multicolumn{3}{c}{LaSOT}&& \multicolumn{3}{c}{LaSOT$_\mathrm{ext}$} && \multicolumn{3}{c}{TrackingNet} && \multicolumn{3}{c}{GOT-10k}\\
    \cmidrule(lr){2-4}
    \cmidrule(lr){6-8}
    \cmidrule(lr){10-12}
    \cmidrule(lr){14-16}
    & AUC&$P$&$P_{\mathrm{Norm}}$ && AUC&$P$&$P_{\mathrm{Norm}}$ && AUC&$P$&$P_{\mathrm{Norm}}$ && AO&$\mathrm{SR_{0.5}}$&$\mathrm{SR_{0.75}}$\\
    \midrule
    OSTrack-B256~\citep{ostrack} &69.1&75.2&78.7 & &47.4&53.3&57.3 & &83.1&82.0&87.8 & &71.0&80.4&68.2\\
    SwinTrack-B384~\citep{swintrack}	&71.3	&76.5	&- & &49.1 &55.6 &- & &84.0	&82.8	&- &	&72.4	&80.5	&67.8\\
    SeqTrack-B256~\citep{seqtrack} &69.9&76.3&79.7&    &49.5&56.3&60.8&    &83.3&82.2&88.3&    &74.7&84.7&71.8\\
    ARTrack-B256~\citep{artrack} &70.4&76.6&79.5&    &46.4&52.3&56.5&     &84.2&83.5&88.7&    &73.5&82.2&70.9\\
    OneTracker-B384~\citep{OneTracker} &70.5&76.5&79.9 & &-&-&- & &83.7&82.7&88.4 & &-&-&- \\
    SUTrack-B224~\citep{sutrack} &73.2&80.5&\underline{{83.4}} & &{53.1}&{60.5}&{64.2} & &{85.7}&85.1&{90.3} & &{77.9}&{87.5}&{78.5}\\
    ARPTrack-B256~\citep{arptrack} &{72.6}&78.5&{81.4} & &52.0&58.7&62.9 & &{85.5}&{85.3}&{90.0} & &{77.7}&{87.3}&{74.3}\\
    RELO-B256~\citep{relo2026} &\underline{{73.3}}&\underline{{80.6}}&\underline{{83.4}} & &\underline{{54.2}}&\underline{{62.2}}&\underline{{65.4}} & &\underline{{86.4}}&\underline{{86.2}}&\underline{{90.7}} & &\textbf{{80.5}}&\textbf{{90.5}}&\underline{{81.2}}  \\
    \midrule
    PATT-B256 (Ours) &\textbf{{75.4}}&\textbf{{84.2}}&\textbf{{85.8}} & &\textbf{{55.4}}&\textbf{{63.6}}&\textbf{{66.8}} & &\textbf{{86.6}}&\textbf{{86.5}}&\textbf{{91.2}} & &\underline{79.7}&\underline{{89.2}}&\textbf{{81.4}}  \\
    \midrule[\heavyrulewidth]
    MixFormer-L320~\citep{mixformer} &70.1&76.3&79.9 & &-&-&- & &83.9&83.1&88.9 & &-&-&-\\
    SimTrack-L224~\citep{simtrack} &70.5&-&79.7 & &-&-&- & &83.4&-&87.4 & &69.8&78.8&66.0\\
    ODTrack-L384~\citep{odtrack} &{74.0}&{82.3}&{84.2} & &53.9&{61.7}&{65.4} & &86.1&{86.7}&{91.0} & &78.2&87.2&77.3 \\
    LoRAT-L224~\citep{LoRAT} &{74.2}&80.9&{83.6} & &52.8&60.0&64.7 & &85.0&84.4&89.5 & &75.7&84.9&75.0\\
    SUTrack-L224~\citep{sutrack} &73.5&80.9&83.3 & &{54.0}&{61.7}&{65.3} & &{86.5}&{86.7}&90.9 & &{81.0}&{90.4}&{82.4}\\
    LoRATv2-L224~\citep{LoRATv2} &74.4&81.2&83.8 & &-&-&- & &85.4&-&- & &76.9&86.3&76.4\\
    RELO-L256~\citep{relo2026} &\underline{{75.1}}&\underline{{83.4}}&\underline{{85.1}} & &\textbf{{57.5}}&\textbf{{66.7}}&\textbf{{69.1}} & &\underline{{87.3}}&\underline{{88.0}}&\underline{{91.6}} & &\underline{{81.8}}&\underline{{91.1}}&\underline{{83.5}} \\
    \midrule
    PATT-L256 (Ours) &\textbf{{76.7}}&\textbf{{85.4}}&\textbf{{87.1}} & &\underline{{56.7}}&\underline{{64.6}}&\underline{{68.2}} & &\textbf{{87.4}}&\textbf{{88.2}}&\textbf{{91.9}} & &\textbf{{82.2}}&\textbf{{91.4}}&\textbf{{84.9}} \\
    \bottomrule
  \end{tabular}
}
\end{table*}

\paragraph{Performance against long-horizon appearance change.}
LaSOT in \Cref{tab:main} is the benchmark most directly aligned with our motivation: as sequences become longer, past templates are more likely to become stale, and the tracker must rely on imperfectly updated target evidence. Compared with temporal baselines such as ARPTrack~\citep{arptrack} which pretrains appearance-motion evolution, and RELO~\citep{relo2026} which adds reward-based localization training and temporal token propagation,
\method changes only the training signal while leaving the deployed tracker unchanged.
This distinction is most visible on LaSOT. Both \method variants  achieve the best overlap and precision metrics within
their scale groups, improving AUC over the strongest same-scale alternative
by $2.1$ points for \method-B256 and $1.6$ points for \method-L256. 
The gains across both AUC and precision suggest that privileged appearance transfer helps the student encode search evidence that remains useful
when historical templates no longer faithfully describe the target.
On LaSOT$_\mathrm{ext}$, the pattern is slightly less uniform:
\method-B256 leads all metrics in the base-scale group, while \method-L256
ranks second to RELO-L256.
Thus, the benefit extends to category shift, although strong online temporal
adaptation remains competitive at larger scale.

\paragraph{Generalization across evaluation regimes.}
On TrackingNet, both \method variants obtain the best results within their scale groups, though the margins are smaller. This indicates that privileged appearance transfer does not harm short-term tracking and can still improve the student representation, even when template staleness is less severe.
The small gains are nevertheless meaningful because ARTrack and ODTrack modify temporal inference through autoregressive trajectory modeling or online token propagation, whereas \method adds no inference-time mechanism beyond the underlying tracker~\citep{artrack,odtrack}.
On GOT-10k, \method-L256 leads all criteria, while \method-B256 shows a more
selective profile: it attains the best strict-overlap success rate
$\mathrm{SR}_{0.75}$, but remains behind RELO-B256 in AO and
$\mathrm{SR}_{0.5}$.
Together, these results suggest that privileged appearance transfer
generalizes beyond long sequences, with its clearest base-scale effect
appearing under stricter localization criteria.

\begin{table}[t]\normalsize
  \centering

  \caption{\textbf{Cross-benchmark transfer on held-out datasets.}
  Results report AUC on TNL2K, NFS, and UAV123, none of which is used for training.}
\label{tab:transfer}
\resizebox{1\linewidth}{!}{
  \setlength{\tabcolsep}{11.25mm}
    \begin{tabular}{lccc}
    \toprule
    Model &TNL2K&NFS&UAV123\\
    \midrule
    TrDiMP~\citep{TMT}&-&66.5&67.5 \\
    TransT~\citep{transt}&50.7&65.7&69.1 \\
    SimTrack~\citep{simtrack}&55.6&-&71.2 \\
    OSTrack~\citep{ostrack}&55.9&66.5&70.7 \\
    SeqTrack-L256~\citep{seqtrack} &56.9&66.9&69.7 \\
    ARTrack-L384~\citep{artrack} &60.3&67.9 &71.2\\
    LoRAT-L224~\citep{LoRAT} &{61.1}&66.0&\textbf{73.3}	\\
    RELO-L256~\citep{relo2026} &\underline{{63.6}}&{{71.3}}&{71.4}	 \\
    \midrule
    PATT-B256 (Ours) &62.9&\underline{71.6}&70.4 \\
    PATT-L256 (Ours) &\textbf{{64.8}}&\textbf{{72.1}}&\underline{72.4} \\
    \bottomrule
    \end{tabular}
  }
\end{table}

\paragraph{Transfer beyond the training benchmarks.}
Table~\ref{tab:transfer} further evaluates transfer on TNL2K, NFS, and UAV123,
none of which is used for training.
\method-L256 achieves the highest AUC on TNL2K and NFS, improving over
the strongest non-\method result by $1.2$ points and $0.8$ points, respectively,
while ranking second on UAV123 with an AUC of $72.4$.
This consistent held-out performance supports the view that privileged
target appearances improve the learned representation rather than merely
fitting the training benchmarks.
The base-scale model is less uniform: \method-B256 is competitive on TNL2K
and ranks second on NFS, but does not improve on UAV123.
The contrast between \method-B256 and \method-L256 suggests that model capacity
may affect how fully the student can absorb training-only privileged targets.
Overall, the results show that \method provides its strongest and most
consistent gains at larger scale.

\subsection{Ablation Studies}
\label{sec:ablation}

We ablate \method-B256 by changing one design choice at a time and report
AUC on LaSOT and LaSOT$_\mathrm{ext}$ in \Cref{tab:ablation}.
The variants test three questions:
whether the gain comes from privileged target appearances,
which form of teacher supervision is most effective,
and how teacher targets should be weighted across samples.

\begin{table}[t]
\centering
\caption{\textbf{Ablation study on LaSOT and LaSOT$_\mathrm{ext}$.}
Results report AUC for \method-B256 and its training variants.
The rows isolate the effects of privileged template information,
representation prediction, and reliability weighting. $\Delta$ denotes the average AUC change relative to the full model across the two benchmarks.}
\label{tab:ablation}
\resizebox{1\linewidth}{!}{
\setlength{\tabcolsep}{9mm}
\begin{tabular}{llccc}
\toprule
\# & Training variant & LaSOT & LaSOT$_\mathrm{ext}$ & $\Delta$\\
\midrule

1 & Full model (\method-B256) &75.4 &55.4 &--\\
\midrule
2 & Student-only baseline &73.0 &52.7 &-2.6\\
3 & Past-only teacher &73.2 &52.6 &-2.5\\
4 & Past+current teacher &75.1 &55.0 &-0.4\\
\midrule
5 & Output imitation &73.5 &53.7 &-1.8\\
6 & Last-layer prediction &74.8 &54.3 &-0.9\\
7 & All-layer prediction &75.5 &55.2 &-0.1\\
8 & Direct matching (no predictor) &73.3 &54.3 &-1.6\\
\midrule
9 &Uniform weighting &75.0 &54.6 &-0.6\\
10 & Quality-only weighting &75.1 &55.2 &-0.3\\
11 & Advantage-only weighting &75.2 &54.8 &-0.4\\

\bottomrule
\end{tabular}
}
\vspace{1.5mm}
\end{table}

\paragraph{Where does the gain originate?}
Rows~$2$-$4$ isolate the information available to the teacher. The student-only baseline and the past-only teacher perform almost identically, showing that a teacher branch alone does not explain the improvement.
Adding the current-frame target crop recovers most of the full model's gain, while the future-frame crop closes the remaining gap. These results indicate that exact current-frame appearance is the dominant source of improvement, while future appearance provides a smaller but  complementary signal about target variation.

\paragraph{What should be transferred?}
Rows~$5$-$8$ examine the form and depth of teacher supervision.
Output imitation improves over the student-only baseline, but remains clearly
below representation prediction, indicating that the teacher's final box does
not capture all useful privileged information.
Predicting representations from four encoder levels outperforms last-layer
prediction, suggesting that privileged appearances shape search evidence at
multiple depths.
Using all layers provides little additional benefit, so a compact set of
representative layers is sufficient.
Directly matching teacher features without a latent predictor also performs poorly:
because the student and teacher receive different template sets, their
representations need not be identical.
The predictor is therefore important because it allows the student to infer the
privileged representation target without forcing exact feature equality.

\paragraph{Which teacher targets should be trusted?}
Rows~$9$-$11$ evaluate how the prediction loss is weighted across samples.
Uniform weighting degrades performance, confirming that privileged inputs do
not make every teacher target equally reliable.
Quality-only weighting discounts poorly localized teachers, while
advantage-only weighting emphasizes samples where the teacher improves over
the student; both recover part of the lost performance.
Combining the two gives the best overall result as it selects teacher
targets that are informative relative to the
student and accurate in absolute terms.

\section{Conclusion and Discussion}
\label{sec:conclusion}

We have introduced \method, a teacher-student framework that uses exact current-
and future-frame target crops as privileged training context for visual
tracking.
Rather than imitating the teacher's final box prediction, the student learns
to predict the teacher's multi-level search representations, with reliability
weighting emphasizing teacher targets that are both accurate and more
informative than the student's own prediction.
After training, the teacher, predictor, reliability computation, and
privileged crops are removed, leaving the standard student-only inference
procedure unchanged.
Across seven benchmarks and two model scales, \method achieves its clearest
gains in long-horizon tracking while remaining competitive in short-term and
cross-benchmark evaluations.

These results also clarify the scope of the present study.
The framework requires frame-level ground-truth bounding boxes to construct privileged
target crops and introduces auxiliary teacher and predictor computation during
training.
It also uses a fixed design for privileged-view selection, replacing two past-frame
templates with one current-frame crop and one future-frame crop.
Thus, the experiments do not yet determine which privileged views are most
informative, how they should be selected across a trajectory, or how much of
the benefit remains when annotation density or training resources are limited.

A natural next step is to make privileged appearance transfer more principled
within the current framework.
Instead of using a fixed current-plus-future template layout, future work could
select privileged views according to information gain, temporal diversity,
occlusion state, teacher uncertainty, or consistency across multiple candidate
crops.
The reliability weight could likewise be extended beyond localization quality
to account for how stable and informative each teacher representation is for
the student.

More broadly, privileged supervision need not be limited to exact target crops.
Dense masks, object parts, depth, optical flow, language descriptions,
long-range reappearance cues, and offline trajectory context could provide
additional training-only signals for learning representations that remain
causal at deployment.
This perspective also connects naturally to emerging practical tracking
settings, including foundation-model-assisted tracking, memory-based trackers,
multi-object tracking, and multi-camera tracking.
In these settings, information from later frames, neighboring objects, or other
cameras may be available during training or offline annotation, even when the
deployed tracker must operate online.
Using such information to train stronger causal trackers is a promising
direction toward robust tracking under appearance change, occlusion,
re-identification, and cross-view ambiguity.

\bibliographystyle{assets/plainnat}
\bibliography{ref}

\end{document}